\documentclass[letterpaper, 10 pt,  journal, twoside]{IEEEtran}  

\IEEEoverridecommandlockouts                              

\usepackage{graphicx} 
\usepackage{amsmath} 
\usepackage{amssymb}  
\usepackage{multirow}

\title{Monocular Semantic Occupancy Grid Mapping with Convolutional Variational Encoder-Decoder Networks
}

\author{Chenyang Lu$^{1}$, Marinus Jacobus Gerardus van de Molengraft$^{2}$, and Gijs Dubbelman$^{1}$
\thanks{$^{1}$Chenyang Lu and Gijs Dubbelman are with the Mobile Perception Systems research cluster of the SPS/VCA group, Dept. of Electrical Engineering,  Eindhoven University of Technology, The Netherlands.
{\tt \footnotesize \{c.lu.2, g.dubbelman\}@tue.nl}}%
\thanks{$^{2}$Marinus Jacobus Gerardus van de Molengraft is with Control System Technology group, Dept. of Mechanical Engineering,  Eindhoven University of Technology, The Netherlands.
{\tt \footnotesize m.j.g.v.d.molengraft@tue.nl}}%
}

\begin{document}

\maketitle
\thispagestyle{empty}
\pagestyle{empty}

\begin{abstract}

In this work, we research and evaluate end-to-end learning of monocular semantic-metric occupancy grid mapping from weak binocular ground truth. The network learns to predict four classes, as well as a camera to bird's eye view mapping. At the core, it utilizes a variational encoder-decoder network that encodes the front-view visual information of the driving scene and subsequently decodes it into a 2-D top-view Cartesian coordinate system. The evaluations on Cityscapes show that the end-to-end learning of semantic-metric occupancy grids outperforms the deterministic mapping approach with flat-plane assumption by more than 12\% mean IoU. Furthermore, we show that the variational sampling with a relatively small embedding vector brings robustness against vehicle dynamic perturbations, and generalizability for unseen KITTI data. Our network achieves real-time inference rates of approx. 35 Hz for an input image with a resolution of 256$\times$512 pixels and an output map with 64$\times$64 occupancy grid cells using a Titan V GPU.
\end{abstract}

\begin{IEEEkeywords}
	 Semantic Scene Understanding; Object Detection, Segmentation and Categorization; Computer Vision for Transportation
\end{IEEEkeywords}

\section{Introduction}

\IEEEPARstart{E}{nvironment} perception is a key task in mobile robot and intelligent vehicle operation. In the past decade, significant progress has been made, mainly due to increased computational power that has unlocked deep learning-based approaches for real-time usage, such as semantic segmentation \cite{Shelhamer2017a, Badrinarayanan2015a, Zhao2016, Chen2016, Meletis2018} and object detection \cite{Girshick2014, Girshick2015, Ren2017, He2017a}. However, it can be argued that, for higher levels of robot and vehicle autonomy, perception and the incorporation of information derived from perception into a consistent world-model, is still a bottleneck. In this work, we therefore research and evaluate the usage of semantic occupancy grid maps, as a means for end-to-end learning of monocular input data to form a world-model.

A world-model typically consists of multiple conceptual layers \cite{Furda2010}, e.g. layers of dynamic objects, permanent static objects, and movable static objects. Furthermore, one can distinguish layers that contain a priori knowledge from the environment, e.g. a global topological map, and layers that are estimated locally while the vehicle is traversing the environment. An occupancy grid map is particularly well-suited to represent the local free-space around the vehicle that is estimated in real-time from sensory input. This is also how we use it and we extend it with three different semantic sub-classes for free-space, namely road, sidewalk, and terrain, besides the usual non free-space class.

A particular branch of deep learning research focuses on convolutional neural networks (CNNs), which have significantly advanced computer vision in the past decade \cite{Krizhevsky2012, Simonyan2014a, He2016a}. At a specific intermediate layer in CNNs, the feature map contains the semantic abstraction of the pixels as well as the inter-pixel 2-D spatial relations between them. The same inter-cell relations also hold for occupancy grids, thereby CNNs are potentially well-suited for end-to-end learning of occupancy grid maps with semantics from image data, which is proposed in this work. We discuss the related work on occupancy grid maps and neural network approaches in more detail in Section II.

\begin{figure}[!tbp]
	\centering
	\includegraphics[width=0.95\linewidth]{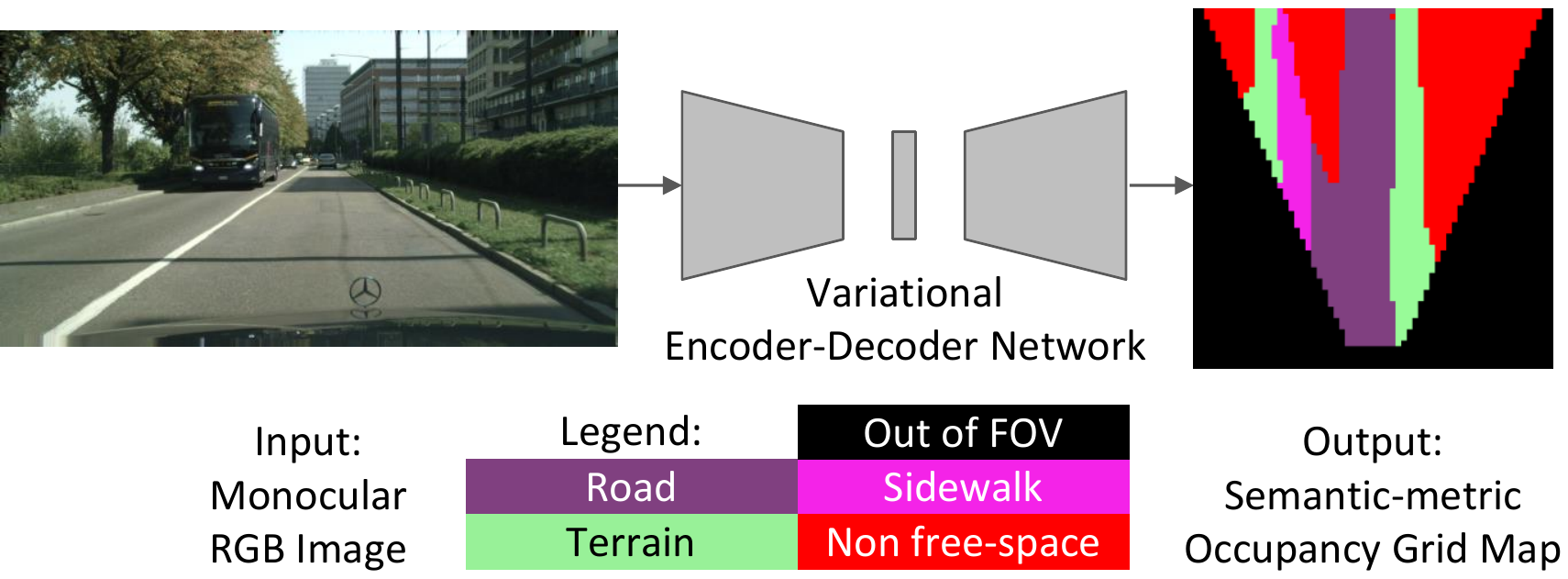}
	\caption{An illustration of the proposed variational encoder-decoder approach. From a single front-view RGB image, our system can predict a 2-D top-view semantic-metric occupancy grid map.}
	\label{example}
\end{figure}

Our approach, which is detailed in Section III, contains the following contributions:
\begin{itemize}
	\item To the best of our knowledge, we are the first to perform end-to-end learning on monocular imagery to produce a semantic-metric occupancy grid map and to achieve real-time inference rates.
	\item We show that this end-to-end monocular approach is intrinsically robust to pitch and roll perturbations, and can generalize on unseen data from different cameras.
	\item We show that, our approach can be trained from weak ground truth and is inherently robust to the sparseness of input data.
\end{itemize}

Considering the above, end-to-end learning of occupancy grids is a promising extension of, or even potentially can partially replace, traditional point-cloud processing techniques. Our approach is evaluated on Cityscapes \cite{Cordts2016a} and KITTI \cite{Geiger2013a} and the details on this are provided in Section IV after which our conclusions are put forward in Section V.

\section{Related work}

The occupancy grid map \cite{Elfes1990} is one of the most popular local metric map representations for mobile robots. Besides range sensors such as RaDAR and LiDAR, occupancy grid maps can also be generated from RGB-D cameras \cite{Himstedt2017a}, stereo vision \cite{Li2014a}, and from fusion of multiple sensors \cite{Oh2016a}. However, the classical occupancy grid maps are without semantics, \textit{i.e.} cells only have two possible states: occupied or not occupied.

More efficient and reliable navigation can be realized if semantics of the environment are utilized. Semantic segmentation is a potential approach to provide additional semantic scene understandings. Most semantic segmentation research has been carried out on RGB images with the goal to estimate a semantic class label for each individual pixel. For this particular task, it can be noted that deep learning methods are surpassing other classical methods in terms of both accuracy and efficiency. One state-of-the-art framework is the fully convolutional network (FCN) \cite{Shelhamer2017a} that utilizes the convolutional feature extractor from other classification networks, such as VGG \cite{Simonyan2014a} or ResNet \cite{He2016a}. Another framework named SegNet \cite{Badrinarayanan2015a}, has a similar structure of auto-encoders. Further research shows that the segmentation quality can be enhanced by applying a conditional random field (CRF) as a post-processing step \cite{Chen2016}. To integrate this in an end-to-end manner, CRFasRNN \cite{Zheng2015} is proposed to form a CRF as a recurrent neural network (RNN) that can be trained directly. Recent research has also performed semantic segmentation in an adversarial manner to produce improved result in terms of segmentation accuracy \cite{Luc2016a}. 

Image-based semantic segmentation methods are usually not directly compatible with vehicle mapping and planning systems. The reason is that in the mainstream state-of-the-art, metric mapping of the environment is deterministically performed in parallel with the semantic segmentation, by which the image-based semantic 3-D mapping is achieved. Sengupta \textit{et al.} \cite{Sengupta2013} project the image semantic labels into 3-D using stereo vision. Based on this, dense pairwise CRFs \cite{Hermans2014, Vineet2015a} and higher order CRFs \cite{Kundu2014, Zhao2016a} are proposed to optimize the 3-D labels. Recently, CNNs are integrated into the mapping pipeline \cite{Yang2017a, McCormac2017, Tateno2017} for a better image segmentation, and furthermore, even depth and pose estimation, which are used in deterministic metric mapping.

Instead of conducting metric mapping and semantic scene understanding separately, our long-term aim is to develop a holistic approach that can estimate metric, semantic, and topological information simultaneously and in real-time. For this we take inspiration from recent work that has shown that deep learning approaches excel in estimating 3-D depth information from monocular \cite{Eigen2014a, Zhou2017, Godard2017} and binocular data \cite{Mayer2016}, which means that the metric information can be learned from photometric data directly. This motivates us to research mapping the environment into  semantic-metric occupancy grid maps directly from monocular input data in an efficient, end-to-end manner with deep neural networks.

\section{Semantic occupancy grid mapping}

In this section, we discuss the details of the aforementioned semantic-metric occupancy grid representation and the detailed structure and training of the proposed deep neural network.
\begin{figure}[!tbp]
	\centering
	\includegraphics[width=\linewidth]{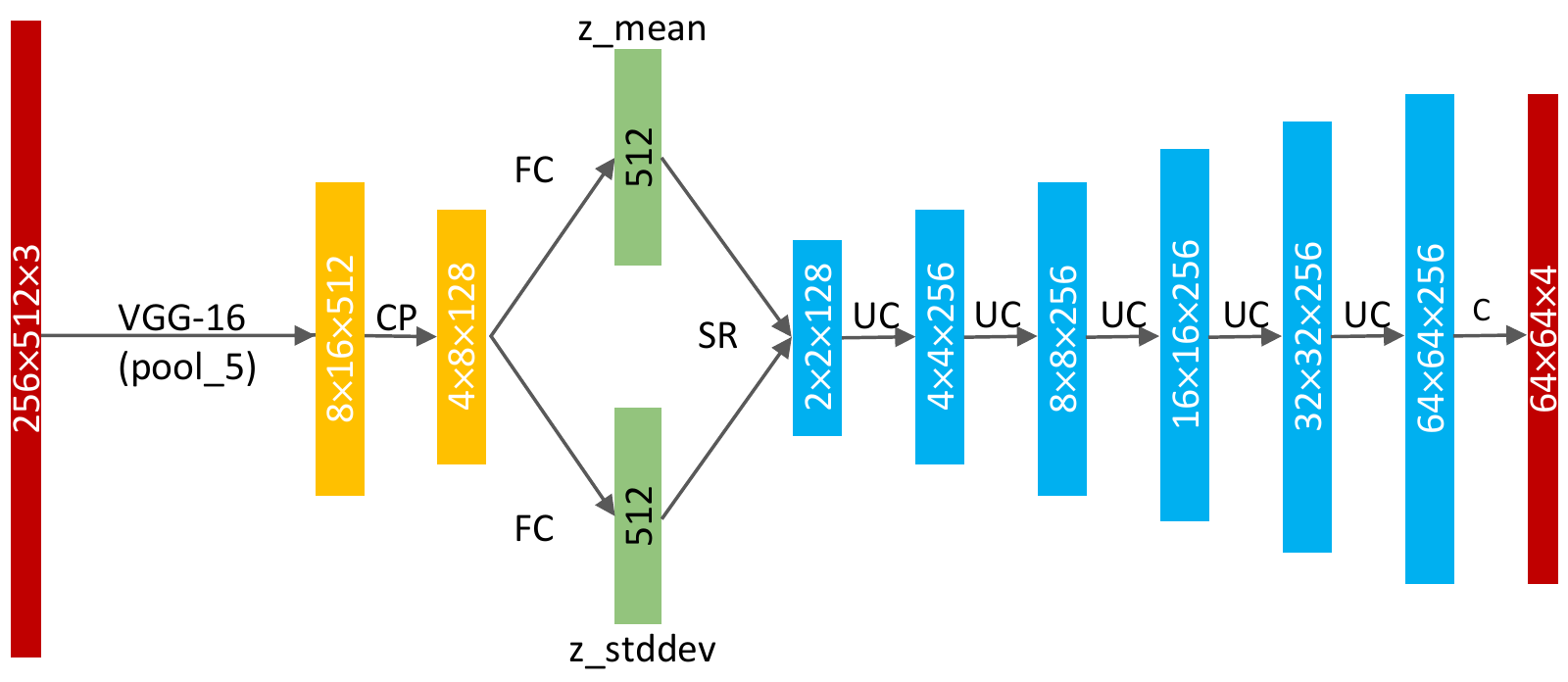}
	\caption{The proposed network structure during the training phase. Every colored block represents a feature map and the arrows between them are neural network layers. Yellow indicates the encoder part while blue indicates the decoder. A pre-trained VGG-16 Net (without fully connected layers) is utilized for feature extraction on top of the input image. Legend: CP = VGG-like convolutional layers (2 layers) with kernel size 3 and 2$\times$2 max pooling, FC = fully connected layer, SR = sample the latent vector with Normal distribution from $z_{mean}$ and $z_{stddev}$ and reshape, UC = one up-convolutional layer and VGG-like convolutional layers (2 layers) with kernel size 3, C = one VGG-like convolutional layer with kernel size 3. Every convolution layer uses batch normalization except the output layer.}
	\label{net}
\end{figure}
\subsection{Map representation}

We extend the classical definition of occupancy grid maps \cite{Elfes1990} to make the map representation contain semantic and metric information as well as suitable for modern deep neural networks.

\textbf{Grid size and perceiving distance:} Sensors mounted on autonomous vehicles such as cameras, RaDARs, and LiDARs usually have a fixed field of view (FOV), and the perception reliability decreases when the perceiving distance increases. To ensure each cell in the grid map, which is represented in 2-D vehicle coordinates, has a reliable status even at large distance, we set each grid map to contain 64$\times$64 cells, with the size of each cell being 0.5$\times$0.5 meters. As the region within 5 meters in front of the vehicle center is never visible, due to the camera's point of view, we apply a 5-meter offset in the grid map w.r.t. the vehicle center.

\textbf{Semantic encoding:} Each cell in the grid map is encoded with one of the following four semantic classes: \textit{road}, \textit{sidewalk}, \textit{terrain}, and \textit{non free-space} (including undetected girds that are behind the foreground objects and out of the camera's FOV). In this configuration, instead of a binary occupancy grid map (free-space or non free-space), the ground area in the map is extended with semantics, which potentially benefits the navigation of mobile robots and autonomous vehicles.

\subsection{Network structure and training}

\begin{figure*}[!tbp]
	\centering
	\includegraphics[width=0.99\linewidth]{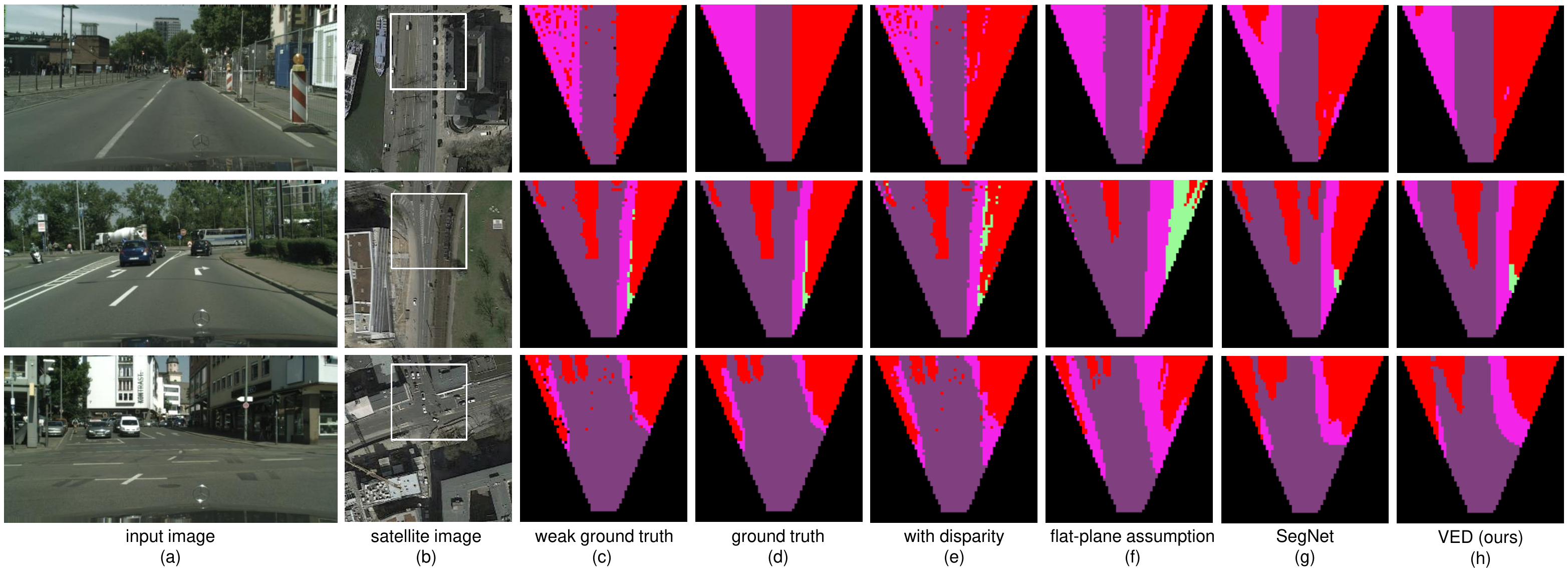}
	\caption{Some visualized mapping examples on the test set with different methods. (a) is the input image from the left RGB camera mounted on the vehicle. (b) is the satellite image corresponding to the RGB image based on the GPS signal from Cityscapes for a better understanding of our work. The region in the white rectangle is focused in the mapping task.  (c) is the weak ground truth map with ground truth semantic segmentation and semi-global matching disparity. 
	(d) is the manually improved ground truth map based on the weak ground truth. (e)-(h) are the mapping result with predicted semantic segmentation and the same disparity, flat-plane assumption geometric transformation, SegNet \cite{Badrinarayanan2015a} baseline method, and our proposed VED method. Grids with black mask are ignored in evaluation as they are out of the camera's FOV or with ignored semantic labels.}
	\label{compare}
\end{figure*}

In this work, instead of implementing a deterministic point cloud based mapping algorithm, we propose an end-to-end learning approach. The proposed system is composed of two components: a low-level feature extractor and a modified version of variational auto-encoder (VAE) \cite{Kingma2013} network on top of the extracted feature map. As in our usage the input and output are not the same, as with a traditional VAE, we refer to our network as a variational encoder-decoder (VED) network. The input of this network is one front-view monocular RGB image, and the output is the top-view occupancy grid map in which each cell is assigned with a semantic class. The network is implemented in PyTorch \cite{Paszke2017} and Figure \ref{net} shows the detailed structure of the network.

\textbf{Feature extractor:} We use a modern CNN model, e.g. VGG-16 \cite{Simonyan2014a}, pre-trained on ImageNet \cite{Russakovsky2015a}, to extract the low level features from the input monocular image. The receptive field of the VGG-16 network is 224$\times$224 pixels. For reasons of efficiency, we use an input resolution of 256$\times$512 pixels. As the receptive field is smaller than the input, the latent features in the output of the VGG-16 network are encoding the semantic information locally instead of on the entire image. This ensures that the spatial information is naturally preserved in the feature map, which is required for decoding the feature map into a top-down view.

\textbf{Training with variational sampling:} The variational auto-encoder \cite{Kingma2013} is originally proposed for learning variational Bayesian models in a neural network fashion. The learned coding vector contains the high-level representation of the input data, which is sampled from a standard normal distribution for later reconstruction. Recent research has shown that, when ground truth for voxel-based learning is incomplete, VAE can be used to produce a reconstruction output that surpasses the ground truth in term of completeness \cite{Schonberger2017}. In our VED case, the ground truth is relatively imprecise (as will be explained in the following subsection), and we aim to mitigate this by using the variational sampling's robustness to imperfect ground truth. In contrast to the VAE model in \cite{Schonberger2017}, several important modifications are made for our VED model: 1) taking the feature map from a modern feature extractor as input, and 2) training in \textit{supervised} encoder-decoder manner instead of an auto-encoder manner.

We denote the encoding probabilistic model as $q_\phi(z|x)$, where $x = f_\gamma(i)$ is the high-level feature from the input image $i$ and $z$ is the latent embedding combined with spatial information and semantics. On top of the encoder, the probabilistic decoder $p_\theta(m|z)$ produces the 2-D grid semantic map $m$ from the latent embedding $z$. The models $f$, $q$, $p$ are organized as neural networks and their parameters $\gamma$, $\phi$, $\theta$ can be learned simultaneously with end-to-end training. As we enforce the latent embedding $z$ to obey the standard normal distribution, the latent loss $\mathcal{L}_{latent}$ is defined as Kullback-Leibler divergence between $z$ and $\mathcal{N}(0,I)$. The mapping loss $\mathcal{L}_{mapping}$ is defined as cross-entropy between the softmax output layer and the one-hot semantic coding of the ground truth. Therefore, the overall loss $\mathcal{L}$ for training is twofold, namely latent loss and mapping loss:
\begin{equation}
\mathcal{L} = \lambda_1 \cdot \mathcal{L}_{latent} + \lambda_2 \cdot \mathcal{L}_{mapping}
\end{equation}
where $\lambda_1$ and $\lambda_2$ are the weights for the balancing of two objectives, which is set as 0.1 and 0.9 respectively in the experiments. We train the network using Adam \cite{kingma2014} optimizer with learning rate $=0.0001$, $\beta_1 = 0.9$, $\beta_2 = 0.999$, and mini-batch sizes of 8 for 60 epochs.

\begin{figure*}[!tbp]
	\centering
	\includegraphics[width=0.9\linewidth]{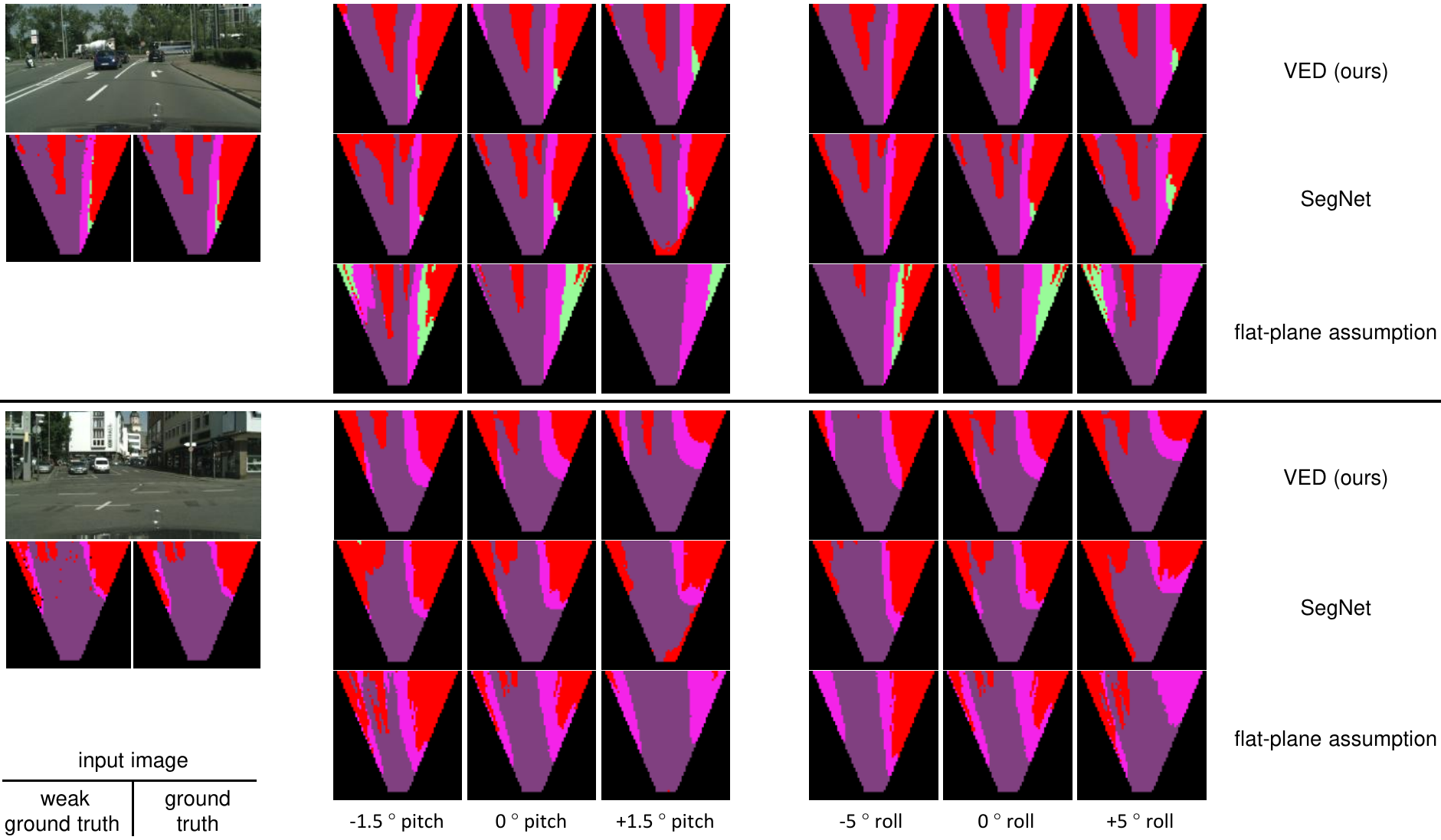}
	\caption{Visualized comparison for different pitch and roll perturbations. We present two examples which are divided by the black horizontal line. For each example, the most left column shows the input RGB image and its corresponding (weak) ground truth. The other columns show the predictions of three monocular approaches in different perturbation settings. }
	\label{disturbance}
\end{figure*}

\textbf{Weak ground truth for training:} One major challenge of our approach is that there is no direct ground truth available, as the top-down view semantic occupancy grid representation is not provided in any publicly available dataset. However, one can utilize datasets that contain front-view image semantic annotations and 3-D information that can be pixel-wised registered as depth/disparity maps. To automatically generate the (weak) ground truth for training, we reconstruct the 3-D point cloud for each frame in vehicle coordinates from the corresponding depth/disparity map, given the intrinsic and extrinsic (the camera's pose in the vehicle coordinates) camera calibration. For each frame of the generated point cloud, given the corresponding front-view image semantic ground truth annotation, a semantic label can be assigned to each 3-D point. Next, we project the 3-D points to the 2-D ground plane and subsequently fill the occupancy grid with pre-defined size. For each cell, a semantic label is assigned, based on the label statics of the cell's points (majority vote). 

The 3-D information registered for the pixels can be noisy (e.g. a disparity map estimated using a stereo matching method) or sparse (e.g. a depth map from LiDAR measurements). It can be argued that the automatically generated ground truth contains noise mainly from the imprecise depth/disparity map, e.g. grid cells can be missed on the road, due to the corresponding depth/disparity region is invalid. For this reason, we refer to the automatically generated ground truth as \textit{weak ground truth}. Some automatically generated weak ground truth examples can be seen in Figure \ref{compare}(c). Note that only for evaluation we have manually annotated 70 top-view grid maps, which is too few for end-to-end training. The ability to train from weak ground truth is an important feature of our neural network based approach.

\section{Experiments}

We conduct the following experiments to demonstrate our approach and to compare its accuracy and robustness with three baseline approaches being: 1) a traditional monocular method that relies on a flat-plane assumption, 2) a traditional binocular approach, and 3) a CNN based approach which is commonly used for segmentation tasks:
\begin{itemize}
	\item \textbf{Quantitative evaluation:} In this experiment, we use the Cityscapes \cite{Cordts2016a} dataset to measure performances employing metrics from semantic image segmentation. 
	\item \textbf{Input disturbance invariance:} We simulate roll and pitch movements of the camera, to investigate the invariance of our approach to such perpetuations.
	\item \textbf{Generalizability to unseen scenarios:} The unseen KITTI \cite{Geiger2013a} dataset containing a semantic domain gap and different camera parameters is used to evaluated the generalizability of the proposed approach.
	\item \textbf{Ablation study:} We compare the results of the proposed network trained with and without variational sampling to better motivate the usage of the variational sampling.
	\item \textbf{Mapping quality invariance w.r.t. resolutions:} We generate maps using two deterministic baseline methods and the proposed neural network method in different resolutions and investigate the advances of the neural network based approach.
	\item \textbf{Semantic latent embedding:} In this small experiment, we research what high-level information is encoded in the latent embedding of our VED approach.
\end{itemize}

\subsection{Dataset and ground truth}

We use the Cityscapes dataset \cite{Cordts2016a} for ground truth generation and experiments, as it provides stereo images with disparity and fine semantic annotations for each pixel. We use the 2975 images in the \textit{training set} for training, and the 500 images in the \textit{validation set} for evaluation and comparison. In our experiments, all the images are resized from 1024$\times$2048 to 256$\times$512 for efficiency. The disparity maps provided from Cityscapes with semi-global matching (SGM) method \cite{Hirschmuller2011} for weak ground truth generation. As discussed in Section III.B, the automatically generated ground truth contains noise. To perform a valid quantitative evaluation, we also manually improved and annotated 70 top-view grid maps in the validation set, based on the visual cue in the corresponding front-view image, which are referred as \textit{ground truth} and visualized in Figure \ref{compare}(d). Furthermore, we use KITTI \cite{Geiger2013a} semantic dataset, which contains 200 images with publicly available semantic annotation, depth maps and camera parameters, to verify the generalizability of different methods. The same procedure is applied for weak ground truth generation and another 70 KITTI samples are manually improved as ground truth for evaluation. 

\subsection{Baseline methods}

Other than our VED approach, there are multiple methods available for mapping sensory data to the proposed map representation. In this paper, we compare our approach with two canonical point cloud based methods and one CNN based SegNet \cite{Badrinarayanan2015a} method:

\subsubsection{Monocular mapping with flat-plane assumption (flat-plane assumption)}

Our first baseline method does not use direct 3-D information, but instead uses a flat-plane assumption to map the output of the semantic segmentation, obtained with a VGG-16 based FCN \cite{Shelhamer2017a} on front-view images, to a top-down view. More precisely, in this method, we assume each pixel in the RGB image which is predicted as one of the ground-like classes (\textit{road}, \textit{sidewalk}, and \textit{terrain}) is located on the ground in 3-D (height being 0 in vehicle coordinates if the vehicle local dynamics is compensated by an IMU \textit{w.r.t.} the static world). With this assumption, the free-space pixels' position in 2-D coordiates can be deterministically solved and mapped using the same method as ground truth generation. However, even given the perfect IMU data, the flat-plane assumption cannot handle scenarios where there is a slope in front of the vehicle, which leads to a map with large offsets. Furthermore, the IMU data can be noisy at each data point, which might further hurt the mapping results, due to the vulnerability of flat-plane assumption. Considering the above reasons and that the data is collected in a relative steady situation, we only use the camera calibration (intrinsic and extrinsic, \textit{i.e.} the camera's pose in vehicle coordinates) instead of IMU data in the experiments. The aim is to outperform this straightforward monocular baseline using our end-to-end learning approach. This method is referred to as \textit{flat-plane assumption} in all figures and tables.

\subsubsection{Binocular mapping (with disparity)}

To provide an upper bound on what we realistically could achieve with our monocular approach, we also validate against a binocular approach. For this baseline, we use the same procedure as for generating the weak ground truth, but the key difference is that now the semantic information is estimated using a VGG-16 based FCN \cite{Shelhamer2017a} instead of the labeled Cityscapes ground truth annotations. This baseline uses binocular image pairs to obtain the corresponding disparity maps for 3-D point cloud generation. In the implementation, the 3-D point clouds are obtained from the Cityscapes disparity maps with SGM method \cite{Hirschmuller2011} and used to fill the occupancy grid. However, note that the disparity maps can also be obtained from other methods, such as stereo network-based approaches \cite{Mayer2016} and monocular network-based approaches \cite{Eigen2014a, Zhou2017, Godard2017}. This method is referred to as \textit{with disparity} in all figures and tables.

\subsubsection{Mapping using canonical CNNs (SegNet)}

We also use a convolutional encoder-decoder based SegNet \cite{Badrinarayanan2015a} to perform the same task. As the size of the input image and output are 256$\times$512 and 64$\times$64, instead of using the original SegNet, we drop the last two decoder modules and applied an additional max-pooling layer with kernel size and stride 1$\times$2 after the last up-pooling layer. We train the SegNet using the same optimizer settings with mini-batch sizes of 2 for 35 epochs. SegNet uses the same VGG-16 \cite{Simonyan2014a} as backbone, and the number of trainable parameters are similar between SegNet ($\approx$28.9M) and the proposed VED network ($\approx$27.5M), which constructs a fair comparison. Note that the SegNet has a significantly larger information bottleneck (8$\times$16$\times$512) than the proposed network (1$\times$512). Hence, it can pass more information to the decoder, which is beneficial for map generation while introducing risks in terms of robustness and generalizability. These will be further discussed in the experimental results. This method is referred to as \textit{SegNet} in all figures and tables.

\begin{table}[!tbp]
	\renewcommand{\arraystretch}{1.5}
	\caption{Quantified performance for different mapping methods.}
	\label{IoU_compare}
	\centering
	\begin{tabular}{c||cc|cc}
		\hline
		\multirow{3}{*}{\bfseries method}		&\multicolumn{2}{c|}{\bfseries weak ground truth} &\multicolumn{2}{c}{\bfseries ground truth} \\
		\cline{2-5}
		\rule{0pt}{20pt}
		& \bfseries \shortstack{mean \\ accuracy} &  \bfseries \shortstack{mean \\ IoU} & \bfseries \shortstack{mean \\ accuracy} &  \bfseries \shortstack{mean \\ IoU} \\
		\hline\hline
		\bfseries with disparity & \textit{91.8} & \textit{80.0} & \textit{85.7} & \textit{70.6}\\
		\hline
		\bfseries flat-plane assumption & 69.9 & 47.2 & 70.1 & 47.2\\
		\hline
		\bfseries SegNet \cite{Badrinarayanan2015a} &  73.7 & \bfseries 61.1 & \bfseries 75.9 & \bfseries 61.5 \\
		\hline
		\bfseries VED (ours) & \bfseries 73.8 & 59.5 &  74.9 &  59.6 \\
		\hline
	\end{tabular}
\end{table}

\subsection{Results}

\begin{table*}[!tbp]
	\renewcommand{\arraystretch}{1.5}
	\caption{Robustness evaluation w.r.t. vehicle local dynamics. The numbers in the brackets indicate the performance downgrade w.r.t. the original performance without perturbation.}
	\label{dynamic_invariance}
	\centering 
	\begin{tabular}{c||cc|cc|cc}
		\cline{2-7}
		&\multicolumn{2}{c|}{\bfseries VED (ours)} &\multicolumn{2}{c|}{\bfseries SegNet \cite{Badrinarayanan2015a}} &\multicolumn{2}{c}{\bfseries flat-plane assumption}\\
		\cline{2-7}
		&\bfseries mean accuracy &  \bfseries mean IoU& \bfseries mean accuracy &  \bfseries mean IoU & \bfseries mean accuracy &  \bfseries mean IoU \\
		\hline\hline
		\bfseries no perturbation &74.9 &59.6 &\bfseries 75.9&\bfseries 61.5 & 70.1 & 47.2\\
		\hline
		\bfseries $\pm$ 1.5$^{\circ}$ pitch &\bfseries72.0 (-2.9)&\bfseries56.2 (-3.4)&69.5 (-6.4) & 54.3 (-7.2) & 53.4 (-16.7) & 35.3 (-11.9)\\
		\hline
		\bfseries $\pm$ 5$^{\circ}$ roll &\bfseries 70.3 (-4.6)&\bfseries 55.1 (-4.5)& 65.7 (-10.2) & 52.0 (-9.5) & 58.3 (-11.8)& 41.5 (-5.7)\\
		\hline
	\end{tabular}
\end{table*}

\subsubsection{Quantitative evaluation}

As our target maps are organized in an image-like fashion, we evaluate the results in terms of mean accuracy and mean intersection-over-union (IoU), averaged over the test samples. The performances of the three mapping methods are provided in Table \ref{IoU_compare}. Note that in this work, the grid cells out of the camera's FOV are used in training but ignored in evaluation and visualization with black mask as they are consistent and trivial for each frame. We report the metrics evaluated on both weak ground truth and manually improved ground truth. The performance of the binocular mapping method (\textit{with disparity}) on weak ground truth is higher than that on manually improved ground truth by a large margin, while the other three methods remain at the same level. This is because the binocular mapping baseline uses exactly the same Cityscapes disparity maps as are also used for weak ground truth generation, which leads to the positive bias when evaluating on the weak ground truth. The aforementioned bias is removed in the metrics evaluated on the manually improved ground truth. In either ground truth setting, it can be seen that the binocular mapping method outperforms the other three monocular methods, as expected. Concerning the monocular methods, the results clearly show that two neural network based methods surpass the flat-plane assumption method by nearly 5\% mean accuracy and 12\% mean IoU. SegNet \cite{Badrinarayanan2015a} provides slightly better performance than our VED network with a margin less than 2\%. This is because that it has a significant larger (8$\times$16$\times$512 instead of 1$\times$512) information bottleneck without prior distribution regularization, which brings small performance improvements but also introducing disadvantages in terms of robustness and generalizability. These disadvantages will be discussed in the following experiments. Given an input with resolution 256$\times$512, our method requires about 28 milliseconds and is thereby able to achieve frame-rates of approx. 35 Hz on a Nvidia Titan V GPU.

\begin{table}[!tbp]
	\renewcommand{\arraystretch}{1.5}
	\caption{Quantified performance for different mapping methods evaluated on the unseen KITTI dataset.}
	\label{KITTI_metrics}
	\centering
	\begin{tabular}{c||cc|cc}
		\hline
		\multirow{3}{*}{\bfseries Method}		&\multicolumn{2}{c|}{\bfseries no perturbation} &\multicolumn{2}{c}{\bfseries $\pm$ 1.5$^{\circ}$ pitch} \\
		\cline{2-5}
		\rule{0pt}{20pt}
		& \bfseries \shortstack{mean \\ accuracy} &  \bfseries \shortstack{mean \\ IoU} & \bfseries \shortstack{mean \\ accuracy} &  \bfseries \shortstack{mean \\ IoU} \\
		\hline\hline
		\bfseries with disparity & \textit{71.1} & \textit{55.9} & \textit{71.1} & \textit{55.9}\\
		\hline
		\bfseries flat-plane assum. & \bfseries 68.3 & \bfseries 51.2 & 52.1(-16.2)& 35.7(-15.5)\\
		\hline
		\bfseries SegNet \cite{Badrinarayanan2015a} &  36.1 & 19.6 & 33.1(-3.0) & 16.2(-3.4) \\
		\hline
		\bfseries VED (ours) & 58.2 & 42.4 & \bfseries 56.1(-2.1) & \bfseries 40.1(-2.3) \\
		\hline
	\end{tabular}
\end{table}

\begin{table}[!tbp]
	\renewcommand{\arraystretch}{1.5}
	\caption{Quantified performance for the proposed VED approach trained with and without variational sampling.}
	\label{ablation}
	\centering
	\begin{tabular}{c||cc|cc}
		\hline
		\multirow{3}{*}{\bfseries Method}		&\multicolumn{2}{c|}{\bfseries no perturbation} &\multicolumn{2}{c}{\bfseries $\pm$ 1.5$^{\circ}$ pitch} \\
		\cline{2-5}
		\rule{0pt}{20pt}
		& \bfseries \shortstack{mean \\ accuracy} &  \bfseries \shortstack{mean \\ IoU} & \bfseries \shortstack{mean \\ accuracy} &  \bfseries \shortstack{mean \\ IoU} \\
		\hline\hline
		\bfseries VED (ours) & \bfseries 74.9 & 59.6 & \bfseries 72.0(-2.9)& \bfseries 56.2(-3.4)\\
		\hline
		\bfseries VED w/o sampling & 74.0 & \bfseries 60.6 & 69.2(-4.8)& 55.0(-5.6) \\
		\hline
	\end{tabular}
\end{table}

\subsubsection{Input disturbance invariance}

While driving, the camera will exhibit roll and pitch perturbations w.r.t. to a stand-still situation. If not accounted for, these perturbations significantly degrade the performance when using a monocular approach based on a flat-plane assumption. Clearly, IMUs can provide orientation information, but the measurement accuracy and time synchronization can be problematic. Ideally, one would want to make the mapping from image coordinates to top-view coordinates intrinsically invariant to such perturbations without using an IMU. We illustrate that our VED network exhibits this invariance. We simulate new input images in the cases of different common orientation disturbances in \textit{pitch} (simulated with vertical pixel offsets) and \textit{roll} (simulated with in-plane rotations around the imaging center) and feed them into different methods. Table \ref{dynamic_invariance} shows the metrics and Figure \ref{disturbance} visualizes some examples with different orientation disturbances. It can be concluded that VED exhibits intrinsic levels of invariance w.r.t. to pitch and roll perturbations, compared to the other monocular baselines. This is mainly because VED network learns to extract the high level semantic-metric information into a low dimensional space, rather than deterministic mapping or direct large feature map passing. Furthermore, it is interesting to note that these results are obtained \textit{without} data augmentation techniques during training that simulate pitch and roll perturbations, which would probably increase the invariance even further.

\subsubsection{Generalizability to unseen scenarios}
\begin{figure*}[!tbp]
	\centering
	\includegraphics[width=\linewidth]{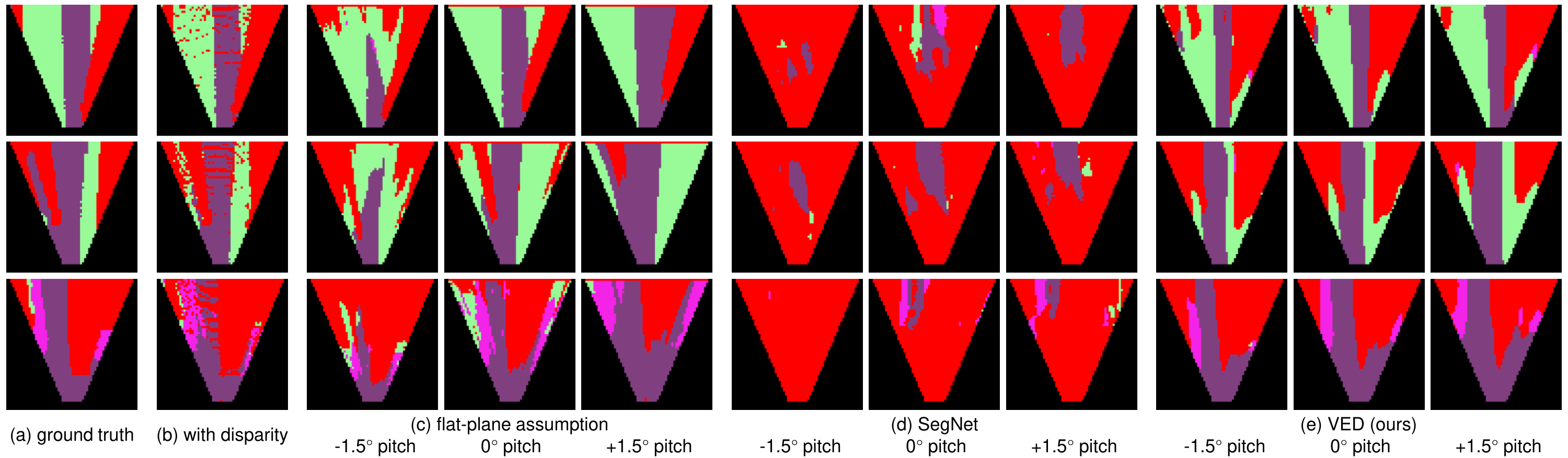}
	\caption{Some visualized mapping examples on the unseen KITTI \cite{Geiger2013a} dataset using different methods. (a) is the manually improved ground truth map based on the weak ground truth. (b) is the deterministic mapping result with predicted semantic segmentation and ground truth depth. (c)-(e) is the mapping result using flat-plane assumption geometric transformation, SegNet \cite{Badrinarayanan2015a} baseline method, and our proposed VED method, with and without pitch perturbations. The black FOV mask is the same as in the Cityscapes, as the input images are cropped for aligning the FOVs of two cameras.}
	\label{compare_kitti}
\end{figure*}
The neural network based methods, \textit{i.e.}, our VED network and SegNet \cite{Badrinarayanan2015a} baseline do not need the camera parameters during testing, which makes the task more challenging when the image is from a different camera. To investigate this, we evaluate our VED network, which is trained on Cityscapes, on the unseen KITTI dataset, which has different camera parameters and semantic domain gaps. We compare the results with the flat-plane assumption (using updated KITTI camera parameters) and the SegNet baseline, see Table \ref{KITTI_metrics}. Obviously, the flat-plane assumption can perform deterministic mapping with the new camera parameters. For the neural network based approaches, in order to make the KITTI data compatible with the networks trained on Cityscapes, we align the horizontal field of view (by cropping KITTI images) and align the vanishing point (by vertical image translation). The results, provided in Table \ref{KITTI_metrics} and Figure \ref{compare_kitti}, show that, although the VED network is trained on camera parameters from Cityscapes \cite{Cordts2016a}, the network can still work with some performance degradation, while the SegNet baseline fails. This degradation (nearly 10\% compared to the flat-plane assumption) is expected, as the camera pose is significantly different between the two datasets. We also observe that without cropping and alignment of the images, the performance of the VED degrades by nearly 13\% for both metrics, which means that these pre-processing steps are necessary when dealing with unseen data. Furthermore, when applying the pitch perturbations on the unseen KITTI dataset, our approach exhibits better performance than the flat-plane assumption, in both degradation and absolute values, which shows the robustness and generalizability of our approach even with changed camera settings and scene domain gaps.

\subsubsection{Ablation study}

To better motivate the usage of variational sampling, we perform an ablation study to investigate the effectiveness of the variational sampling in our mapping task. We train the proposed VED network with only one modification: the embedding vector is directly passed from the encoder to the decoder, instead of randomly sampled and regularized based on the outputs of two fully-connected layers. Table \ref{ablation}, shows the performance of the VED network with and without the variational sampling. Two networks exhibit similar performance when no perturbation is applied. However, the usage of variational sampling improve the robustness against perturbations: the performance degradation with sampling is about 2\% less than that without sampling.

\subsubsection{Mapping quality invariance}

\begin{figure}[!tbp]
	\centering
	\includegraphics[width=0.98\linewidth]{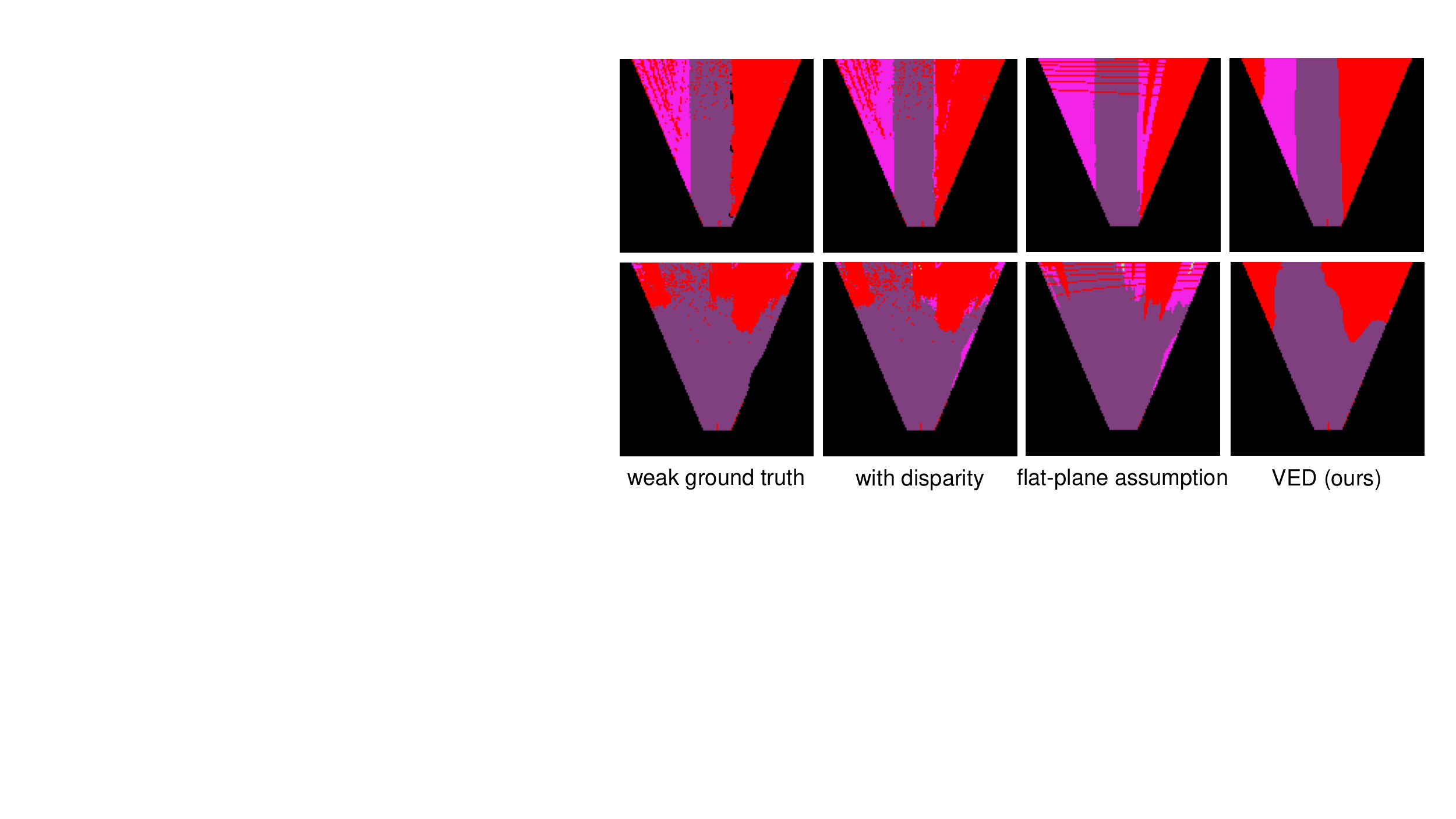}
	\caption{Examples of weak ground truth map and predictions from different mapping approaches in high resolution (128$\times$128 pixels) setting. Both point cloud based approaches produce maps with certain artifact patterns, while the VED network produces maps with acceptable quality.}
	\label{high_resolution}
\end{figure}

In our experiments, the resolution of the map representation is set to be 64$\times$64 pixels, while it can be extended to any other resolution, such as 128$\times$128 pixels or even higher. With the output resolution increasing, the side effects will appear in point cloud based mapping approaches: the artifacts will exhibit because the points registered for the grid at far distance are insufficient for a reliable majority vote. In Figure \ref{high_resolution}, we show some  prediction examples using deterministic approaches and our VED approach with the map resolution being 128$\times$128 pixels. It can be observed that at large distance, semantic information is lost in some grids with certain patterns in point cloud based methods, which degrades mapping quality, while the network based method will not exhibit this behavior. Our approach is intrinsically invariant to point cloud density as we extract high level semantic-metric information from images directly and achieve higher map resolution with up-convolution operations. 

\subsubsection{Semantic latent embedding}

\begin{figure}[!tbp]
	\centering
	\includegraphics[width=\linewidth]{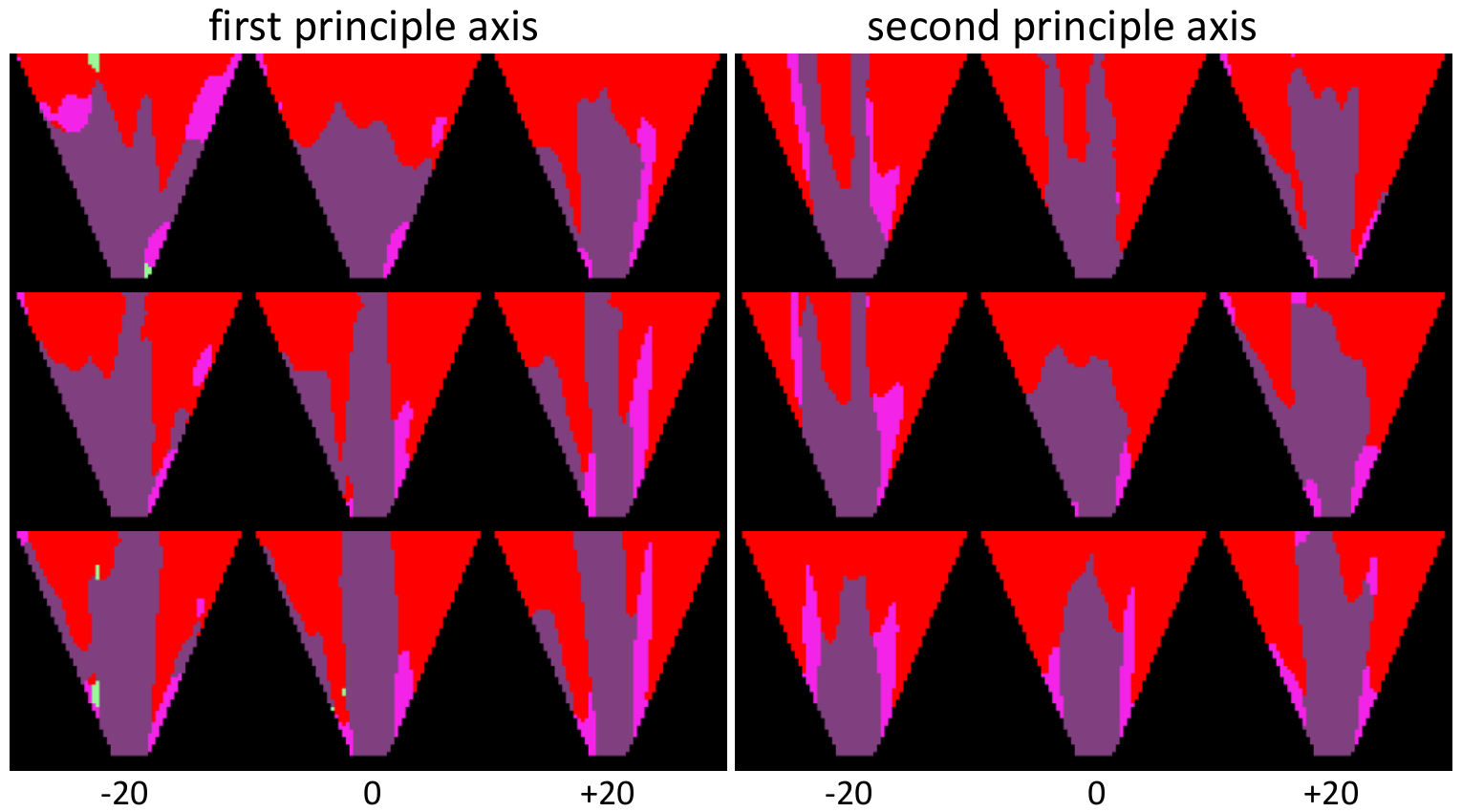}
	\caption{PCA perturbation analysis. The numbers are indicating the perturbation values applied on the first and second principle axis.}
	\label{pca}
\end{figure}

The latent representaion in our proposed network is supposed to encode both high-level semantic and spatial information into an embedding vector with 512 dimensions. As our system handles complicated data in real urban environments, some attributes in the vector might be highly correlated, which makes it difficult to perform direct attribute analysis. To analyze the effectiveness of our encoding and decoding system separately, we conduct the principal component analysis (PCA) on 500 test images' embedding vectors. We apply perturbations on the first and second principal axises and visualize the modified map predictions, which are illustrated in Figure \ref{pca}. It can be noted that the first principal axis is mainly encoding the width of the drivable space in front of the vehicle, and the second one is encoding the size of the non free-space area near the center of FOV. This shows that our network indeed learns to encode semantic and spatial understanding from monocular image into a latent embedding vector. As mentioned earlier this spatial understanding provides the network with robustness to pitch and roll perturbations as well allows for up-sampling the resolution of the occupancy grid map.

\section{Conclusion}

In this work, we proposed a novel real-time neural network based end-to-end mapping system, which requires a single front-view image from a monocular camera and from it estimates a top-view semantic-metric occupancy grid map. It is shown that our VED approach outperforms a monocular system using a flat-plane assumption, and exhibits better robustness and generalizability, due to its variational sampling over a relatively low dimensional embedding space, when compared to the canonical neural network baseline. We have verified that the network can learn semantics as well as metric spatial information, by investigating the latent embedding that it uses. Although further research is certainly required to bring our VED approach to a level at which it can be applied, our work demonstrates that, occupancy grids, although already several decades old, are still a very relevant and powerful representation and that they link very well with state-of-the-art methods from deep learning, which can enhance or even partially replace traditional point cloud processing techniques. In future work, we aim to further leverage on deep learning and predict the road layout beyond the camera's FOV.





\bibliographystyle{./bib/IEEEtran}
\bibliography{./bib/library}

\begin{thebibliography}{10}
\providecommand{\url}[1]{#1}
\csname url@samestyle\endcsname
\providecommand{\newblock}{\relax}
\providecommand{\bibinfo}[2]{#2}
\providecommand{\BIBentrySTDinterwordspacing}{\spaceskip=0pt\relax}
\providecommand{\BIBentryALTinterwordstretchfactor}{4}
\providecommand{\BIBentryALTinterwordspacing}{\spaceskip=\fontdimen2\font plus
\BIBentryALTinterwordstretchfactor\fontdimen3\font minus
  \fontdimen4\font\relax}
\providecommand{\BIBforeignlanguage}[2]{{%
\expandafter\ifx\csname l@#1\endcsname\relax
\typeout{** WARNING: IEEEtran.bst: No hyphenation pattern has been}%
\typeout{** loaded for the language `#1'. Using the pattern for}%
\typeout{** the default language instead.}%
\else
\language=\csname l@#1\endcsname
\fi
#2}}
\providecommand{\BIBdecl}{\relax}
\BIBdecl

\bibitem{Shelhamer2017a}
E.~Shelhamer, J.~Long, and T.~Darrell, ``{Fully Convolutional Networks for
  Semantic Segmentation},'' \emph{IEEE TPAMI}, vol.~39, no.~4, pp. 640--651,
  2017.

\bibitem{Badrinarayanan2015a}
V.~Badrinarayanan, A.~Kendall, and R.~Cipolla, ``{SegNet: A Deep Convolutional
  Encoder-Decoder Architecture for Image Segmentation},'' \emph{IEEE TPAMI},
  vol.~39, no.~12, pp. 2481--2495, 2017.

\bibitem{Zhao2016}
H.~Zhao, J.~Shi, X.~Qi, X.~Wang, and J.~Jia, ``{Pyramid Scene Parsing
  Network},'' in \emph{CVPR}, 2017, pp. 6230--6239.

\bibitem{Chen2016}
L.-C. Chen, G.~Papandreou, I.~Kokkinos, K.~Murphy, and A.~L. Yuille,
  ``{DeepLab: Semantic Image Segmentation with Deep Convolutional Nets, Atrous
  Convolution, and Fully Connected CRFs},'' \emph{IEEE TPAMI}, vol.~40, no.~4,
  pp. 834--848, 2018.

\bibitem{Meletis2018}
P.~Meletis and G.~Dubbelman, ``{Training of Convolutional Networks on Multiple
  Heterogeneous Datasets for Street Scene Semantic Segmentation},'' in
  \emph{IV}, 2018.

\bibitem{Girshick2014}
R.~Girshick, J.~Donahue, T.~Darrell, and J.~Malik, ``{Rich Feature Hierarchies
  for Accurate Object Detection and Semantic Segmentation},'' in \emph{CVPR},
  2014, pp. 580--587.

\bibitem{Girshick2015}
R.~Girshick, ``{Fast R-CNN},'' in \emph{ICCV}, 2015, pp. 1440--1448.

\bibitem{Ren2017}
S.~Ren, K.~He, R.~Girshick, and J.~Sun, ``{Faster R-CNN: Towards Real-Time
  Object Detection with Region Proposal Networks},'' \emph{IEEE TPAMI},
  vol.~39, no.~6, pp. 1137--1149, 2017.

\bibitem{He2017a}
K.~He, G.~Gkioxari, P.~Dollar, and R.~Girshick, ``{Mask R-CNN},'' in
  \emph{ICCV}, 2017, pp. 2980--2988.

\bibitem{Furda2010}
A.~Furda and L.~Vlacic, ``{An object-oriented design of a World Model for
  autonomous city vehicles},'' in \emph{IV}, 2010, pp. 1054--1059.

\bibitem{Krizhevsky2012}
A.~Krizhevsky, I.~Sutskever, and G.~E. Hinton, ``{ImageNet Classification with
  Deep Convolutional Neural Networks},'' in \emph{NIPS}, 2012.

\bibitem{Simonyan2014a}
K.~Simonyan and A.~Zisserman, ``{Very Deep Convolutional Networks for
  Large-Scale Image Recognition},'' \emph{CoRR}, vol. abs/1409.1, 2014.

\bibitem{He2016a}
K.~He, X.~Zhang, S.~Ren, and J.~Sun, ``{Deep Residual Learning for Image
  Recognition},'' in \emph{CVPR}, 2016, pp. 770--778.

\bibitem{Cordts2016a}
M.~Cordts, M.~Omran, S.~Ramos, T.~Rehfeld, M.~Enzweiler, R.~Benenson,
  U.~Franke, S.~Roth, and B.~Schiele, ``{The Cityscapes Dataset for Semantic
  Urban Scene Understanding},'' in \emph{CVPR}, 2016, pp. 3213--3223.

\bibitem{Geiger2013a}
A.~Geiger, P.~Lenz, C.~Stiller, and R.~Urtasun, ``{Vision meets robotics: The
  KITTI dataset},'' \emph{IJRR}, vol.~32, no.~11, pp. 1231--1237, 2013.

\bibitem{Elfes1990}
A.~Elfes, ``{Occupancy Grids: A Stochastic Spatial Representation for Active
  Robot Perception},'' in \emph{Proceedings of the Sixth Conference on
  Uncertainty in Artificial Intelligence}, 1990, pp. 136--146.

\bibitem{Himstedt2017a}
M.~Himstedt and E.~Maehle, ``{Online semantic mapping of logistic environments
  using RGB-D cameras},'' \emph{IJIJARS}, vol.~14, no.~4, pp. 1--13, 2017.

\bibitem{Li2014a}
Y.~Li and Y.~Ruichek, ``{Occupancy grid mapping in urban environments from a
  moving on-board stereo-vision system},'' \emph{Sensors (Switzerland)},
  vol.~14, no.~6, pp. 10\,454--10\,478, 2014.

\bibitem{Oh2016a}
S.-I. Oh and H.-B. Kang, ``{Fast Occupancy Grid Filtering Using Grid Cell
  Clusters From LIDAR and Stereo Vision Sensor Data},'' \emph{IEEE Sensors
  Journal}, vol.~16, no.~19, pp. 7258--7266, 2016.

\bibitem{Zheng2015}
S.~Zheng, S.~Jayasumana, B.~Romera-Paredes, V.~Vineet, Z.~Su, D.~Du, C.~Huang,
  and P.~H.~S. Torr, ``{Conditional Random Fields as Recurrent Neural
  Networks},'' in \emph{ICCV}, 2015, pp. 1529--1537.

\bibitem{Luc2016a}
P.~Luc, C.~Couprie, S.~Chintala, and J.~Verbeek, ``{Semantic Segmentation using
  Adversarial Networks},'' \emph{arXiv preprint, arXiv:1611.08408}, 2016.

\bibitem{Sengupta2013}
S.~Sengupta, E.~Greveson, A.~Shahrokni, and P.~H.~S. Torr, ``{Urban 3D semantic
  modelling using stereo vision},'' in \emph{ICRA}, 2013, pp. 580--585.

\bibitem{Hermans2014}
A.~Hermans, G.~Floros, and B.~Leibe, ``{Dense 3D semantic mapping of indoor
  scenes from RGB-D images},'' in \emph{ICRA}, 2014, pp. 2631--2638.

\bibitem{Vineet2015a}
V.~Vineet, O.~Miksik, M.~Lidegaard, M.~Niebner, S.~Golodetz, V.~A. Prisacariu,
  O.~Kahler, D.~W. Murray, S.~Izadi, P.~Peerez, and P.~H.~S. Torr,
  ``{Incremental dense semantic stereo fusion for large-scale semantic scene
  reconstruction},'' in \emph{ICRA}, 2015, pp. 75--82.

\bibitem{Kundu2014}
A.~Kundu, Y.~Li, F.~Dellaert, F.~Li, and J.~M. Rehg, ``{Joint semantic
  segmentation and 3D reconstruction from monocular video},'' in \emph{ECCV},
  2014, pp. 703--718.

\bibitem{Zhao2016a}
Z.~Zhao and X.~Chen, ``{Building 3D semantic maps for mobile robots using RGB-D
  camera},'' \emph{Intelligent Service Robotics}, vol.~9, no.~4, pp. 297--309,
  2016.

\bibitem{Yang2017a}
S.~Yang, Y.~Huang, and S.~Scherer, ``{Semantic 3D occupancy mapping through
  efficient high order CRFs},'' in \emph{IROS}, 2017, pp. 590--597.

\bibitem{McCormac2017}
J.~McCormac, A.~Handa, A.~Davison, and S.~Leutenegger, ``{SemanticFusion: Dense
  3D semantic mapping with convolutional neural networks},'' in \emph{ICRA},
  2017, pp. 4628--4635.

\bibitem{Tateno2017}
K.~Tateno, F.~Tombari, I.~Laina, and N.~Navab, ``{CNN-SLAM: Real-Time Dense
  Monocular SLAM with Learned Depth Prediction},'' in \emph{CVPR}, 2017, pp.
  6565--6574.

\bibitem{Eigen2014a}
D.~Eigen, C.~Puhrsch, and R.~Fergus, ``{Depth Map Prediction from a Single
  Image using a Multi-Scale Deep Network},'' \emph{arXiv preprint,
  arXiv:1406.2283}, 2014.

\bibitem{Zhou2017}
T.~Zhou, M.~Brown, N.~Snavely, and D.~G. Lowe, ``{Unsupervised Learning of
  Depth and Ego-Motion from Video},'' in \emph{CVPR}, 2017, pp. 6612--6619.

\bibitem{Godard2017}
C.~Godard, O.~M. Aodha, and G.~J. Brostow, ``{Unsupervised Monocular Depth
  Estimation with Left-Right Consistency},'' in \emph{CVPR}, 2017, pp.
  6602--6611.

\bibitem{Mayer2016}
N.~Mayer, E.~Ilg, P.~Hausser, P.~Fischer, D.~Cremers, A.~Dosovitskiy, and
  T.~Brox, ``{A Large Dataset to Train Convolutional Networks for Disparity,
  Optical Flow, and Scene Flow Estimation},'' in \emph{CVPR}, 2016, pp.
  4040--4048.

\bibitem{Kingma2013}
D.~P. Kingma and M.~Welling, ``{Auto-Encoding Variational Bayes},'' \emph{arXiv
  preprint, arXiv:1312.6114}, 2013.

\bibitem{Paszke2017}
A.~Paszke, G.~Chanan, Z.~Lin, S.~Gross, E.~Yang, L.~Antiga, and Z.~Devito,
  ``{Automatic differentiation in PyTorch},'' in \emph{NIPS Workshop}, 2017.

\bibitem{Russakovsky2015a}
O.~Russakovsky, J.~Deng, H.~Su, J.~Krause, S.~Satheesh, S.~Ma, Z.~Huang,
  A.~Karpathy, A.~Khosla, M.~Bernstein, A.~C. Berg, and L.~Fei-Fei, ``{ImageNet
  Large Scale Visual Recognition Challenge},'' \emph{IJCV}, vol. 115, no.~3,
  pp. 211--252, 2015.

\bibitem{Schonberger2017}
J.~L. Sch{\"{o}}nberger, M.~Pollefeys, A.~Geiger, and T.~Sattler, ``{Semantic
  Visual Localization},'' in \emph{CVPR}, 2018.

\bibitem{kingma2014}
D.~Kingma and J.~Ba, ``{Adam: A method for stochastic optimization},''
  \emph{arXiv preprint, arXiv:1412.6980}, 2014.

\bibitem{Hirschmuller2011}
H.~Hirschmuller, ``{Stereo Processing by Semiglobal Matching and Mutual
  Information},'' \emph{IEEE TPAMI}, vol.~30, no.~2, pp. 328--341, 2008.

\end{thebibliography}

\end{document}